\documentclass[conference]{IEEEtran}
\IEEEoverridecommandlockouts
\usepackage{cite}
\usepackage{amsmath,amssymb,amsfonts}
\usepackage{algorithmic}
\usepackage{algorithm}
\usepackage{graphicx}
\usepackage[table]{xcolor}
\usepackage{textcomp}
\usepackage{subcaption}
\usepackage{makecell}
\usepackage{xcolor}
\usepackage{tikz}
\usetikzlibrary{arrows, positioning}
\def\BibTeX{{\rm B\kern-.05em{\sc i\kern-.025em b}\kern-.08em
    T\kern-.1667em\lower.7ex\hbox{E}\kern-.125emX}}
\begin{document}

\title{EEschematic: Multimodal-LLM Based AI Agent for Schematic Generation of Analog Circuit\\

\thanks{Chang Liu is sponsored by Peter Denyer's PhD Scholarship at The University of Edinburgh}
}

\author{
\IEEEauthorblockN{Chang Liu}
\IEEEauthorblockA{\textit{School of Engineering} \\
\textit{The University of Edinburgh}\\
Edinburgh, UK \\
C.Liu-134@sms.ed.ac.uk}
\and
\IEEEauthorblockN{Danial Chitnis}
\IEEEauthorblockA{\textit{School of Engineering} \\
\textit{The University of Edinburgh}\\
Edinburgh, UK \\
d.chitnis@ed.ac.uk}
}

\maketitle

\begin{abstract}
Circuit schematics play a crucial role in analog integrated circuit design, serving as the primary medium for human understanding and verification of circuit functionality. While recent large language model (LLM)-based approaches have shown promise in circuit topology generation and device sizing, most rely solely on textual representations such as SPICE netlists, which lack visual interpretability for circuit designers. 

To address this limitation, we propose EEschematic, an AI agent for automatic analog schematic generation based on a Multimodal Large Language Model (MLLM). EEschematic integrates textual, visual, and symbolic modalities to translate SPICE netlists into schematic diagrams represented in a human-editable format. The framework uses six analog substructure examples for few-shot placement and a Visual Chain-of-Thought (VCoT) strategy to iteratively refine placement and wiring, enhancing schematic clarity and symmetry. Experimental results on representative analog circuits, including a CMOS inverter, a five-transistor operational transconductance amplifier (5T-OTA), and a telescopic cascode amplifier, demonstrate that EEschematic produces schematics with high visual quality and structural correctness. 

\end{abstract}


\section{Introduction}

Machine learning based methods have significantly accelerated research in analog circuit design, particularly in topology generation and sizing. For example, research \cite{zhao2022analog} develop an analog Predefined Building Block Library (PBBL) for analog sub-circuits and leverages Deep Reinforcement Learning (DRL) to generate building block sequences and decode as Python code to select topologies. AnalogCoder uses LLMs to generate Python code for PySpice, developing a feedback loop to enrich the circuit tool library to improve LLMs' ability in analog design \cite{lai2025analogcoder}. AmpAgent\cite{liu2024ampagent} employs a Retrieval-Augmented Generation (RAG) method, manually converting the circuit images from research papers to SPICE netlists to enhance LLM's knowledge of analog topology. Most approaches represent circuits as SPICE netlists and Python code, describing connectivity in a textual format. However, analog circuit designers typically rely on schematic diagrams to understand signal paths and circuit topologies, rather than textual descriptions. Consequently, even when ML models generate optimal circuit netlists, they are often complex for designers to interpret and verify, and converting them into schematics is time-consuming and prone to mistakes \cite{matsuo2024schemato}. Therefore, an automatic approach for converting analog circuit netlists to visual schematics is desired.

Research on automated netlist-to-schematic generation is mainly focused on digital circuits. Existing synthesis and implementation tools such as Vivado \cite{feist2012vivado} and Synopsys Design Compiler \cite{compiler2016synopsys} support automated design and visualization. However, these gate-level approaches are not well-suited for transistor-level analog schematic generation, as analog schematics emphasise factors including symmetry and visual clarity, rather than sequential gate connections. Few studies have explored analog schematic generation. Early works used symmetry-based placement \cite{arsintescu1996method} and net-crossing minimization \cite{lee1992aesthetic}, but these methods failed to generalize across circuit types. RL-based approaches generate schematics using building block classification, successfully handling a few analog circuits, yet still rely on comprehensive libraries of subcircuits \cite{hsu2022automatic}. Schemato used domain-specific LLMs to convert netlists to schematics based solely on textual information, achieving a 76\% compilation success rate on their dataset. However, it struggles to generate schematics for circuits with more than five components with accurate connectivity \cite{matsuo2024schemato}.

Recently, Multi-modal Large Language Models (MLLMs) such as Gemini~2, GPT~4, and Claude~4 have demonstrated outstanding abilities in integrating textual and visual understanding and reasoning. MLLMs allow flexible modelling of relationships across modalities, which suggests they have the potential to generate schematics by integrating multimodal information, such as textual descriptions, circuit diagrams, and symbolic netlists, into coherent circuit representations.

In this work, we propose EEschematic, a multimodal AI agent for automatic analog schematic generation. We reduce reliance on schematic libraries by introducing a few basic analog substrcuture examples for a few-shot guidance during the initial placement stage. Additionally, we integrate a Visual Chain-of-Thought (VCoT) prompting for placement and wiring, enabling iterative visual reasoning over schematic images and JSON representations. Experimental results on different analog circuits, including the 5T-OTA and the telescopic cascode amplifier, demonstrate that EEschematic can generate visually clear and structurally correct schematics.

\section{Methodology}

\subsection{Circuit Representations}
\begin{figure}[t]
    \centering
    \includegraphics[width=0.48\textwidth]{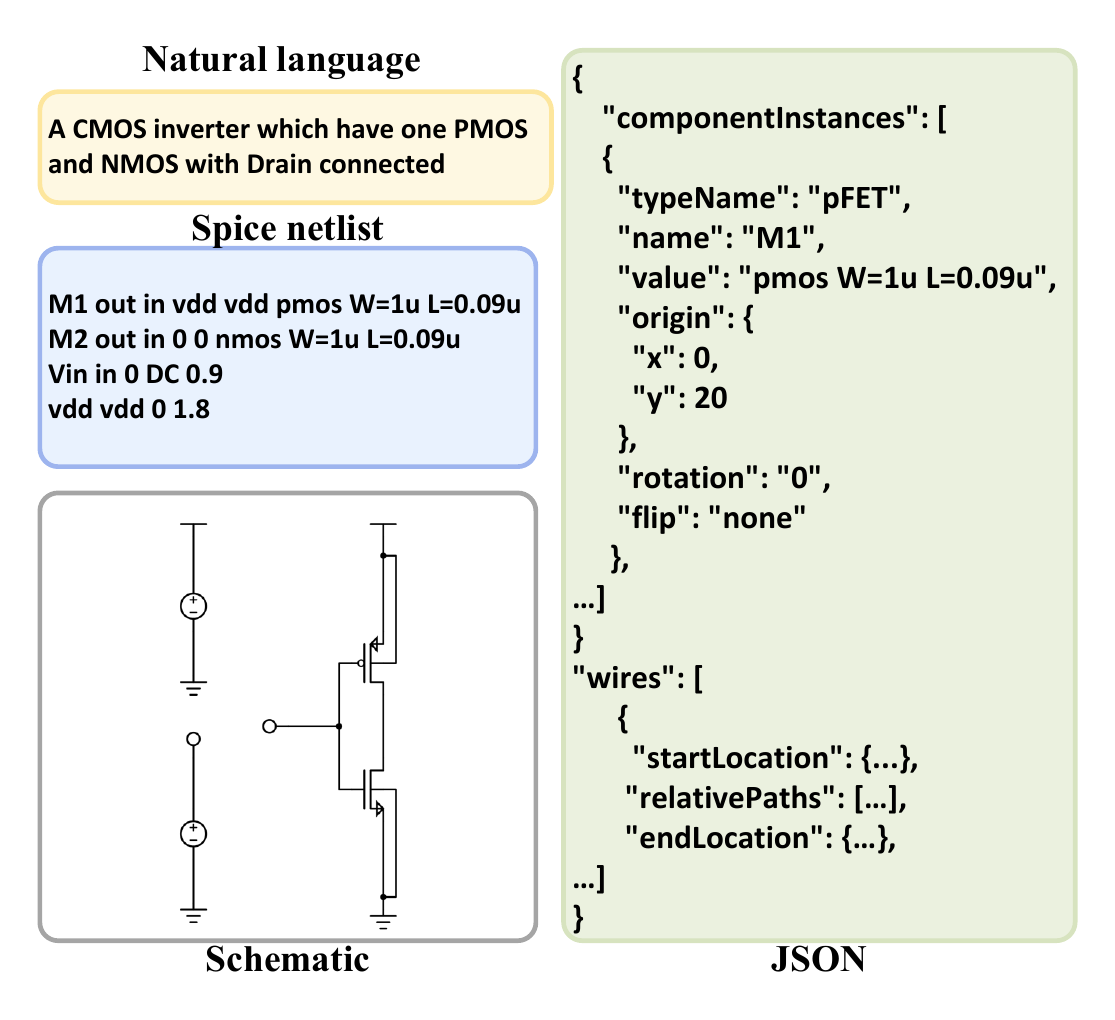}
    \caption{Different modalities of circuit representation, including natural language description, SPICE netlist, circuit schematic diagram, and corresponding position and wiring information within a JSON schema.}
    \label{fig:circuit_represent}
\end{figure}
In this work, circuit schematic generation is formulated as a multimodal task that involves constructing a natural language description, an SPICE netlist, a circuit schematic diagram, and corresponding position and wiring information within a JSON file for the schematic renderer, as shown in Fig.~\ref{fig:circuit_represent}. 

The SPICE netlist encodes the electrical connectivity and component parameters of the circuit in a machine-readable format. The natural language description provides a human-understandable explanation of circuit functionality, summarizing the roles of key components and their interactions. The schematic diagram offers a visual representation of the circuit topology, enabling intuitive analysis and refinement. The position and wiring data specify the spatial information and interconnections, ensuring consistent schematic rendering. Together, these modalities establish a shared context that enables MLLMs to associate textual, symbolic, and geometric information for more accurate schematic generation.

\subsection{Sub-circuit Examples}
\definecolor{lightblue}{RGB}{233,242,254}
\renewcommand{\arraystretch}{1.2}
\begin{table}[t]
\centering
\caption{Sub-circuit examples predefined as LLM's context for schematic generation. The JSON file is represented in images, which are generated from the schematic renderer to show the spatial structure of the circuits.}
\label{tab:examples set}
\begin{tabular}{c|c|c}
\hline
\rowcolor{lightblue}
& \textbf{Single Cascode} & \textbf{Single Current Source} \\ 
\hline\hline

\makecell{JSON \\ schema \\ represented \\ schematic}
& \makecell{\centering\includegraphics[width=0.07\linewidth]{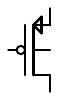}}
& \makecell{\centering\includegraphics[width=0.07\linewidth]{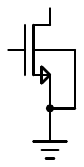}} \\ 
\hline

\makecell{Natural \\ Language \\ Description}
& \makecell{Single transistor. \\ The source \\ and drain are connected...}
& \makecell{Single transistor. \\ The source should \\ be connected to...} \\ 
\hline

\rowcolor{lightblue}
& \textbf{Diode Connected} & \textbf{Two-Transistor Cascode} \\ 
\hline\hline

\makecell{JSON \\ schema  \\ represented \\ schematic}
& \makecell{\centering\includegraphics[width=0.07\linewidth]{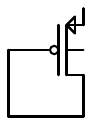}}
& \makecell{\centering\includegraphics[width=0.15\linewidth]{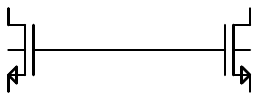}} \\ 
\hline

\makecell{Natural \\ Language \\ Description}
& \makecell{Single transistor \\ with gate and drain \\ shorted together...}
& \makecell{Two transistors with \\ gates connected together. \\ The left one ...} \\ 
\hline

\rowcolor{lightblue}
& \textbf{Differential Pair} & \textbf{Current Mirror} \\ 
\hline\hline

\makecell{JSON \\ schema \\ represented \\ schematic}
& \makecell{\centering\includegraphics[width=0.14\linewidth]{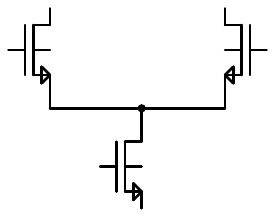}}
& \makecell{\centering\includegraphics[width=0.14\linewidth]{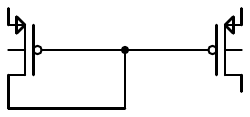}} \\ 
\hline

\makecell{Natural \\ Language \\ Description}
& \makecell{Three transistors \\ with two sources \\ connected to ...}
& \makecell{Two PMOS transistors \\ with gates \\ connected together....} \\ 
\hline

\hline
\end{tabular}
\end{table}
LLMs' ability to reason from few-shot examples makes it possible to learn the underlying principles of analog structure from a limited set of examples and generalize this understanding to generate correct placement, demonstrating that it is not necessary to maintain a comprehensive example set containing every possible structure variant, such as both PMOS and NMOS-based implementations. In this work, we manually constructed a reduced analog substructure example set based on the building block library \cite{zhao2022analog} to store instructions and examples for the placement process. 

This set comprises three key collections: First, substructure placement descriptions, which serve as placement guidelines. Second, the corresponding SPICE netlists form the textual foundation for both structural recognition and simulation-driven reasoning by the LLM. Third, a JSON schema encodes positional information, which provides examples for reference. An overview of all the examples is presented in Table~\ref{tab:examples set}. 

\subsection{Initial Placement and Wiring}

The initial placement aims at establishing a foundational structure that ensures proper symmetry, flipping, and rotation of components. The LLM-guided process begins by analyzing the SPICE netlist to identify circuit substructures, with examples provided as context for the LLM. The LLM first analyzes the netlist to identify substructures, then determines their spatial relationships and orientations to generate an initial schematic placement that serves as the basis for subsequent refinement.
The initial wiring between components is generated automatically by a custom-defined wiring algorithm, which reconstructs circuit connectivity from the SPICE netlist and node location data, automatically identifying and mapping electrical connections between device terminals. Terminals sharing the same net name are considered connected and grouped accordingly, forming the foundation for a hierarchical connection graph. The connection priority is assigned in the following order by their connection complexity: VDD, GND, Gate, Drain, Source, Bulk, and finally I/O, resistor, capacitor, and other pins. In the end, A post-processing step removes conflicting connections, yielding start–end coordinate pairs that serve as the foundation for schematic visualization and further geometric refinement.

\subsection{Optimization}

\begin{figure}[t]
    \centering
    \includegraphics[width=0.5\textwidth]{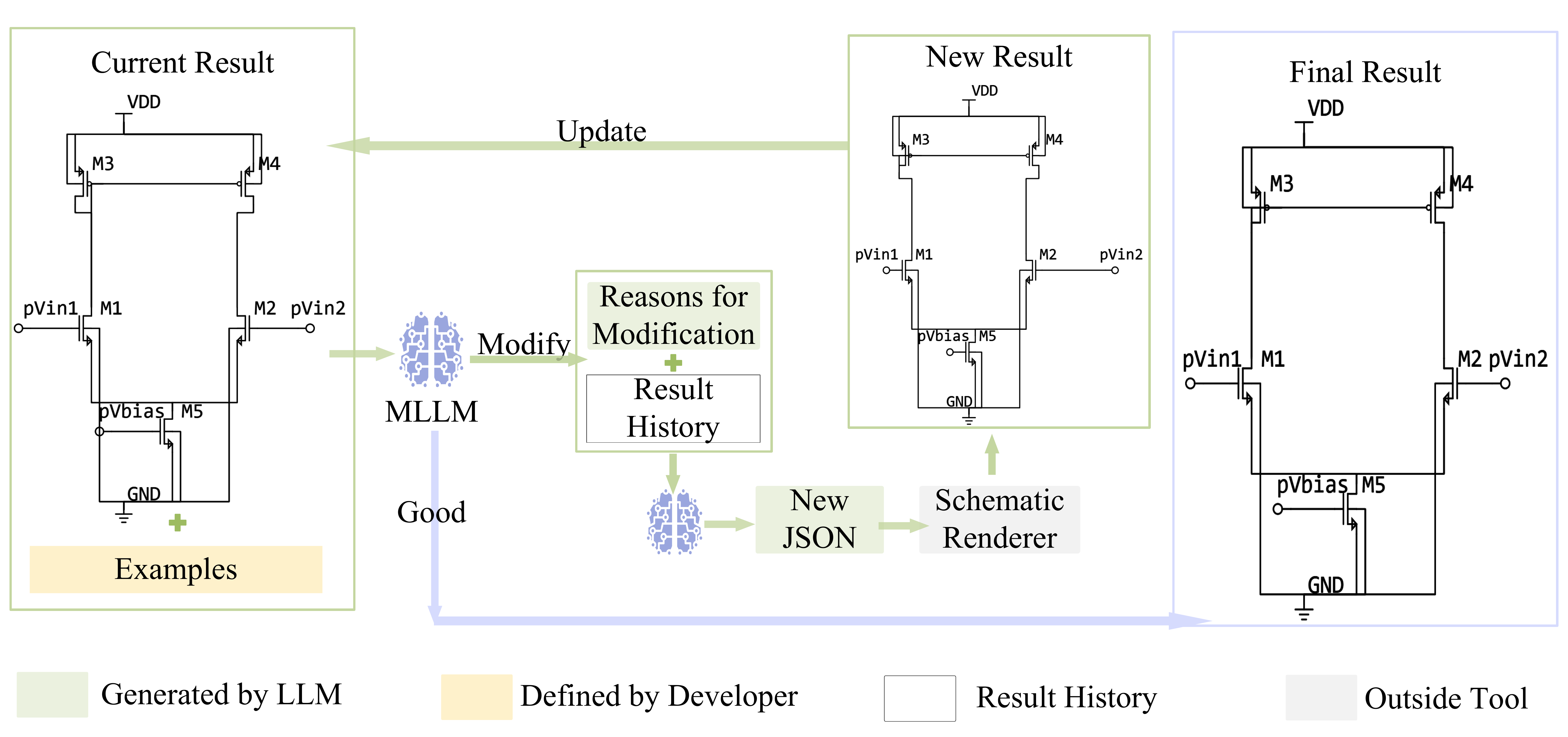}
    \caption{The inner optimization loop for placement. The MLLM compares the current schematic with reference examples to determine whether modification is needed. If required, the model generates reasoning chains to guide schematic refinement. The updated schematic and its corresponding JSON data are then iteratively fed back into the inner optimization loop, enabling continuous improvement through multimodal reasoning.}
    \label{fig:place_loop}
\end{figure}

In this work, the optimization of the initially generated schematic is divided into two stages: placement and wiring optimization. During these stages, Visual Chain-of-Thought (VCoT) prompting\cite{shao2024visual} is employed to construct an inner reasoning loop that guides the MLLM through multimodal information, thereby enhancing both the clarity and aesthetic quality of the schematic.

To implement VCoT in the placement optimization, a two-step interaction is established between EEschematic and the MLLM API. In the first step, the MLLM receives the current schematic in image format and determines whether modifications are necessary by comparing it against reference examples in its context. These reference examples include both good and bad examples: a poor placement case, illustrating misplaced devices and disorganized wiring, and a preferred placement case, demonstrating the desired spatial organization and clean connectivity pattern.

The MLLM evaluates the similarity between the current schematic and the reference examples to decide whether modifications are required. If the schematic is identified as suboptimal by the LLM, the VCoT prompt then guides the MLLM to generate reasoning chains that explain the necessary modifications. The model subsequently uses the result history, integrating schematic images and corresponding JSON data, as in-context information to produce an improved placement. This iterative reasoning process enables in-context learning, allowing the MLLM to better understand the relationship between spatial arrangement, wiring complexity, and schematic readability. Each newly generated result is then passed to the schematic renderer, and the outcomes are sent back for modification. The previous result is incorporated into the result history, forming a self-improving inner optimization loop. The overall process of the proposed VCoT-based placement optimization is illustrated in Fig.~\ref{fig:place_loop}.
After placement optimization, the modified schematic is passed to the next stage for detailed wiring optimization, which follows a similar VCoT-guided methodology.

\section{Results}

\begin{table}[t]
\centering
\caption{Correctness and aesthetics for different circuits. In this experiment, a maximum of 10 iterations is assigned to placement and 20 for wiring optimization, with 10 trials performed per circuit.}
\resizebox{0.5\textwidth}{!}{
\begin{tabular}{c|c|c|c|c}
\hline
\textbf{Circuit} & \textbf{Correctness} & \textbf{Aesthetics} & \makecell{Avg. Iter. \\ for Placement.} & \makecell{Avg. Iter. \\ for Wiring.}  \\ \hline
\hline
Inverter & 9/10 & 9/10 & 3 & 3\\ 
5T-OTA & 9/10 & 8/10 & 3 & 4 \\ 
Telescope cascode & 9/10 & 5/10 & 4 & 6 \\ 
\hline
\end{tabular}
}
\label{tab:result}
\end{table}

\begin{figure*}[t]
    \centering
    \includegraphics[width=0.95\textwidth]{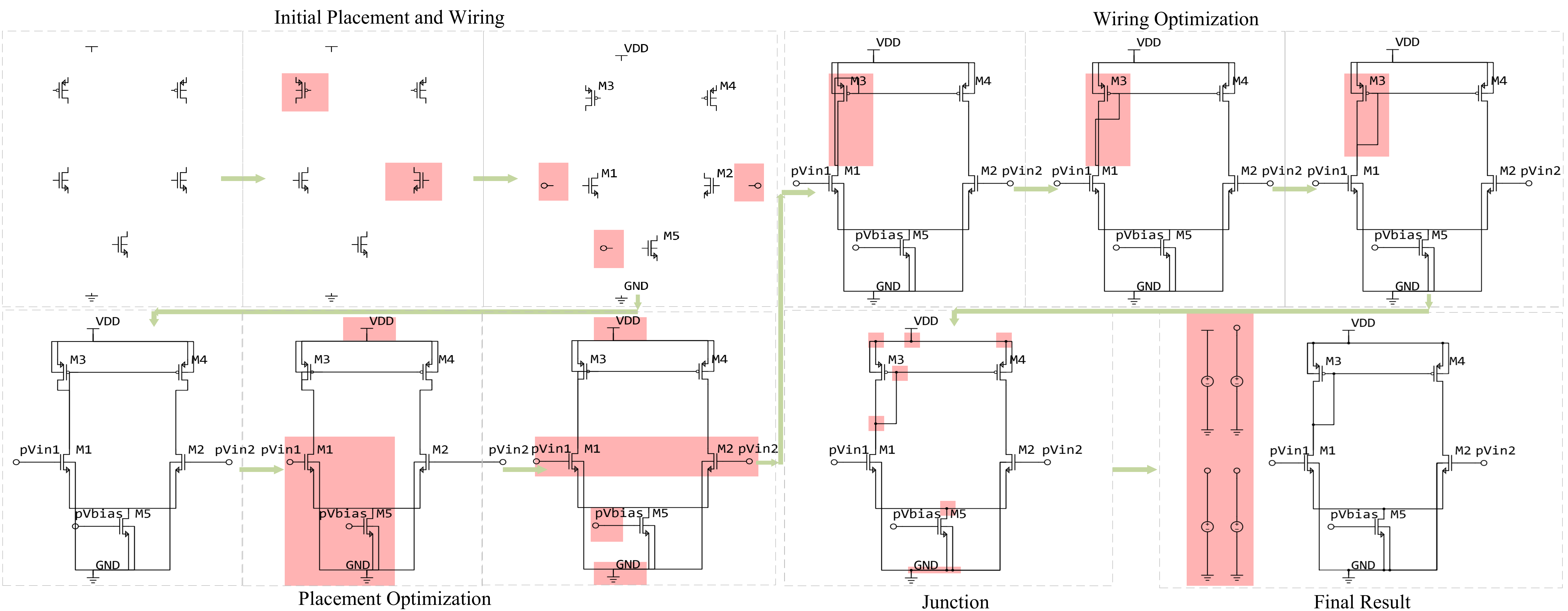}
    \caption{The complete process of schematic generation for the 5T-OTA from initial placement to the final result. Changes from previous iterations are highlighted in red blocks.}
    \label{fig:result_whole}
\end{figure*}

\begin{figure}[t]
    \centering
    \begin{minipage}[t]{0.18\textwidth}
        \centering
        \includegraphics[width=\linewidth]{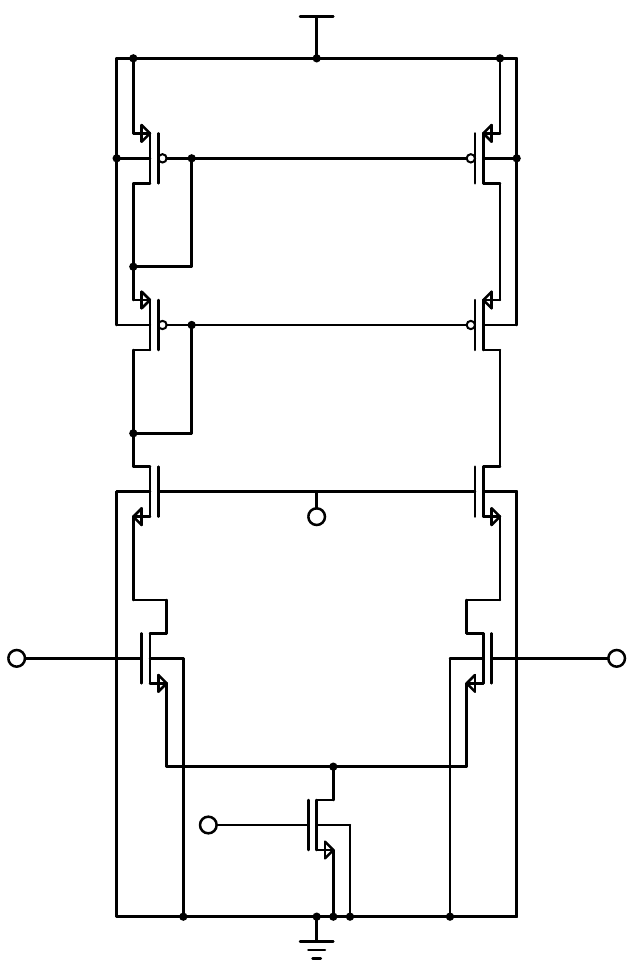}
        \caption*{(a) Best result}
        \label{fig:tel_best}
    \end{minipage}%
    \hfill
    \begin{minipage}[t]{0.19\textwidth}
        \centering
        \includegraphics[width=\linewidth]{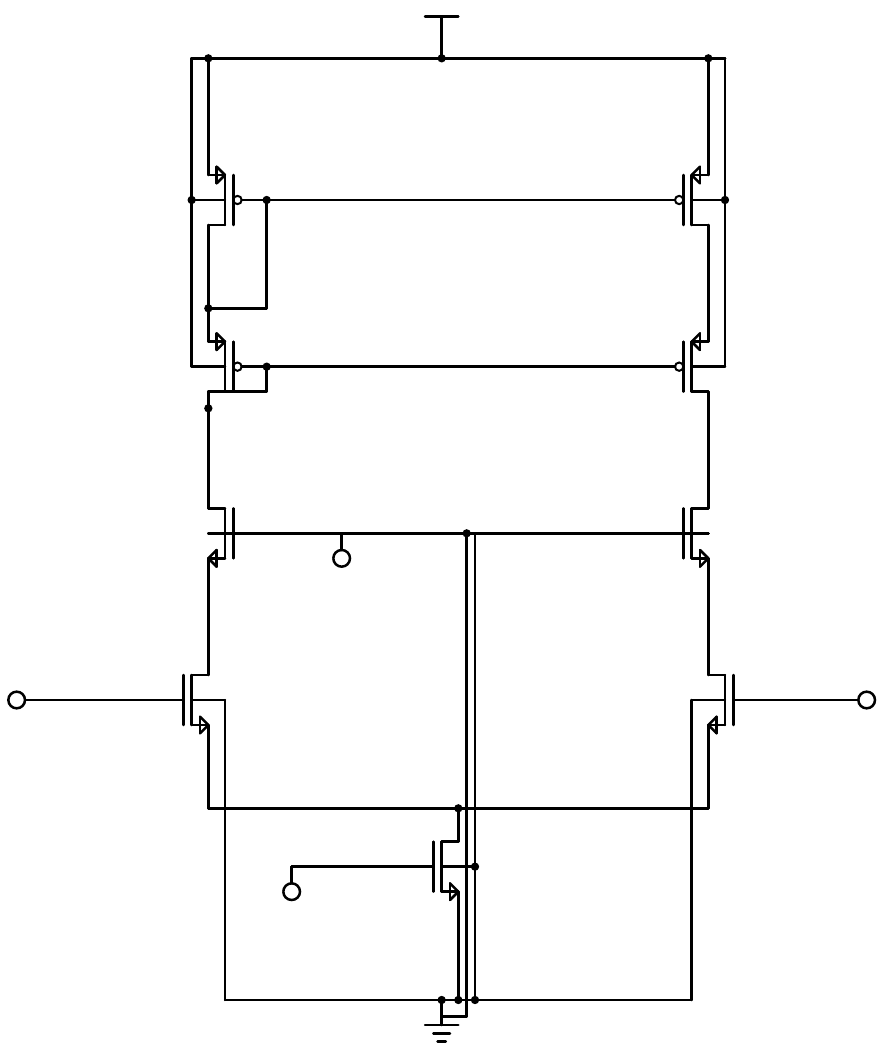}
        \caption*{(b) Worst result}
        \label{fig:tel_worst}
    \end{minipage}%
    \hfill
    \begin{minipage}[t]{0.18\textwidth}
        \centering
        \includegraphics[width=\linewidth]{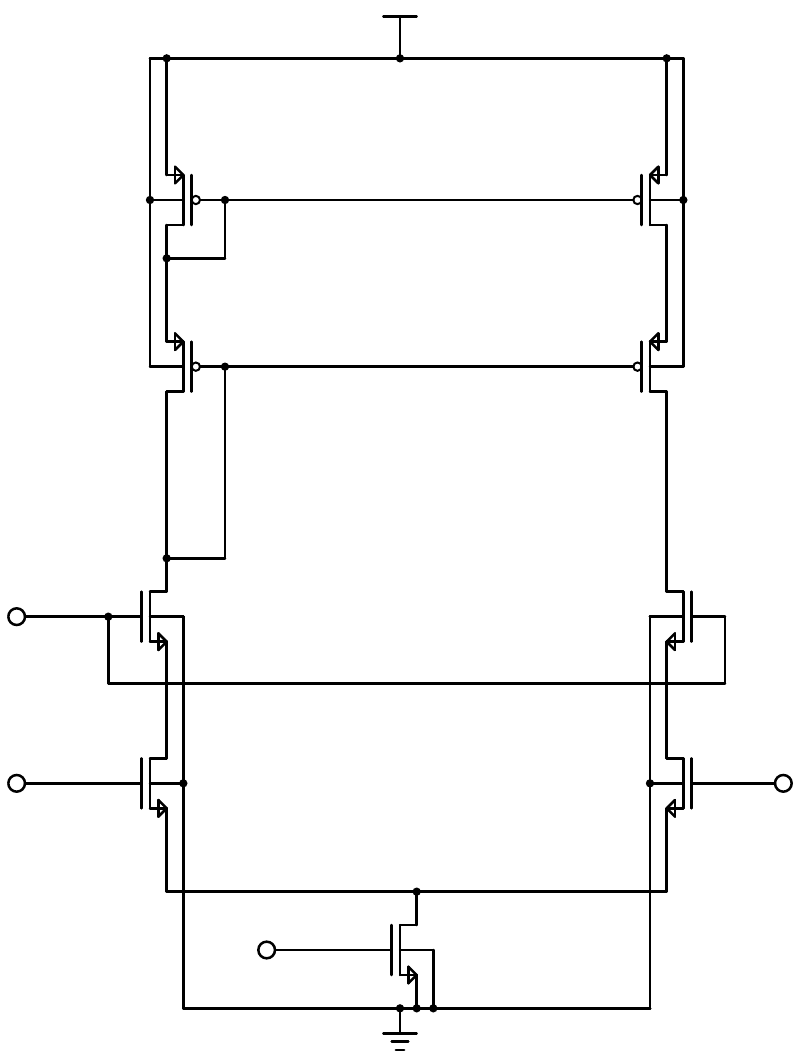}
        \caption*{(c) Aesthetic result}
        \label{fig:tel_middle}
    \end{minipage}
    \hfill
    \begin{minipage}[t]{0.20\textwidth}
        \centering
        \includegraphics[width=\linewidth]{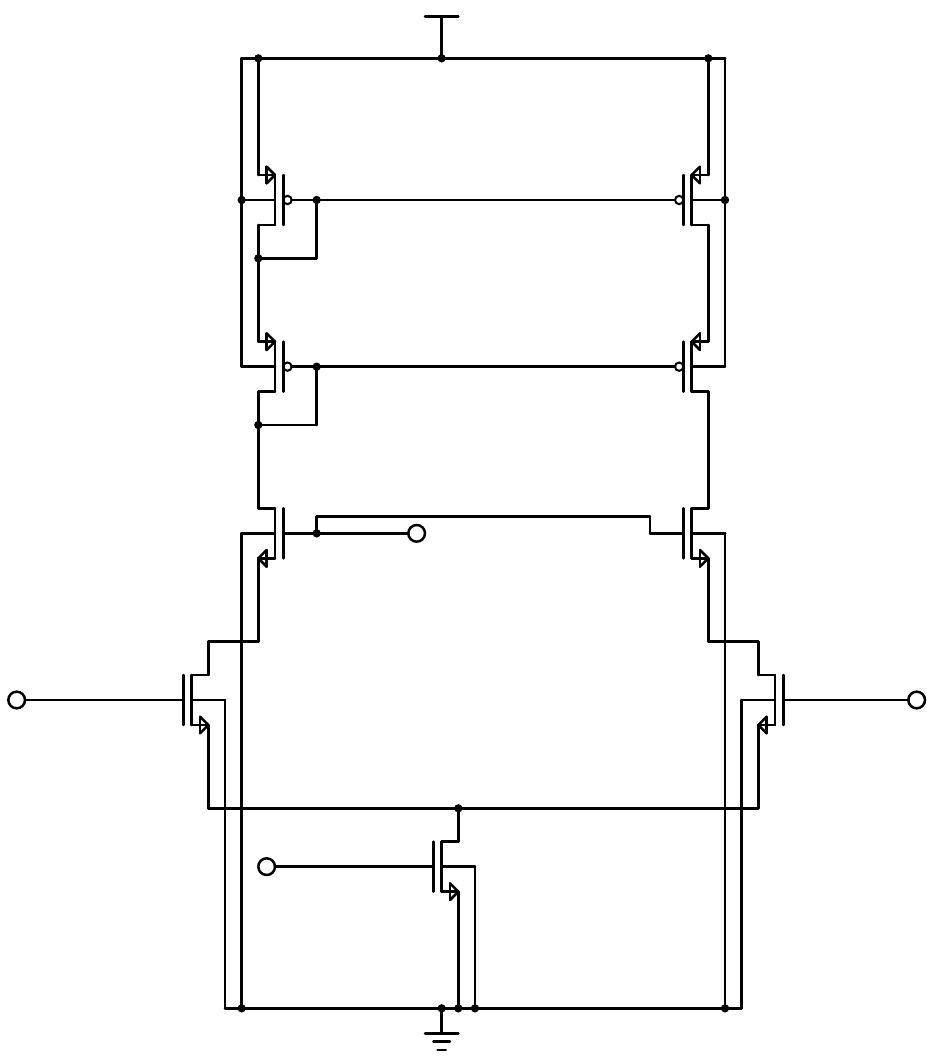}
        \caption*{(d) Correct result}
        \label{fig:tel_middle}
    \end{minipage}
    \caption{Visualization of the JSON-based placement results of a telescopic cascode amplifier over 10 trials: (a) the best result, (b) the worst result, (c) an example of a correct and aesthetic result. (d) an example of a correct but not aesthetic result.}
    \label{fig:tel_result}
\end{figure}

We conducted experiments on three representative analog circuits: an inverter, a 5T-OTA, and a telescopic cascode amplifier, using Gemini 2.0 Flash as the evaluation MLLM. To evaluate the performance of EEschematic, we introduced visual correctness and aesthetics as the key evaluation metrics. Correctness represents the structural validity of the schematic by verifying device connectivity and ensuring no overlaps occur in components and wiring. In contrast, aesthetics are evaluated subjectively by manually assessing visual quality for symmetry, alignment, wire compactness, and overall clarity. 

The results are summarized in Table~\ref{tab:result}. Simpler circuits, such as the inverter, achieve both high correctness and strong aesthetics with minimal iterations. In contrast, the telescopic cascode amplifier, with denser connectivity and more complex spatial relationships, maintains high correctness but exhibits lower aesthetic scores, reflecting a more challenging visually optimized schematics within limited iterations.

Fig.~\ref{fig:result_whole} illustrates the schematic generation process for the 5T-OTA. Starting from the netlist, the model performs initial placement and then refines alignment and connectivity through three placement and three wiring iterations, ultimately producing a coherent schematic. This demonstrates the model’s multimodal reasoning capability to iteratively optimize based on JSON schema and visual information given.

Fig.~\ref{fig:tel_result} compares the best, worst, an example of a correct and aesthetic result and an example of a correct but not aesthetic result of the telescopic cascode amplifier. The best case shows strong symmetry and compact wiring, while the worst case contains incorrect and asymmetric connections. This highlights both the potential and variability of the LLM-based placement optimization process.

\section{Discussion}
This work found that the proposed EEschematic, a MLLM-based AI Agent, can effectively generate circuit schematics from input netlists for different circuit types. However, some limitations remain.
Firstly, this work focused primarily on fundamental analog circuits. While these serve as representative test cases for validating schematic reasoning and spatial organization, they only serve as a single block in more advanced circuit topologies, such as multi-stage amplifiers and mixed-signal systems. Future studies should extend the ability of these complex circuits to further assess the scalability and generalization of the proposed approach.
Secondly, the subjective evaluation of the final schematic aesthetics introduces a degree of human bias. Future work could incorporate quantitative visual metrics or expert-validated benchmarks to ensure a more objective assessment.
Thirdly, despite increasing the number of optimization iterations, a few generated schematics still exhibit structural or connectivity errors. Incorporating constraint-guided refinement strategies could help improve reliability and ensure convergence toward fully correct schematic representations. 

\section{Conclusion}
We have developed EEschematic, an MLLM-based AI agent that automates the transformation from netlist to schematic. Using a few substructure examples for in-context learning and Visual Chain-of-Thought (VCoT) prompts to link textual and visual reasoning, the agent learns spatial organization and connectivity patterns, producing coherent and interpretable schematics.
The proposed agent successfully generated schematics for three basic circuits: an inverter, a 5T-OTA, and a telescopic cascode amplifier, achieving a 90\% correctness rate within an average of 10 iterations. The generated results exhibit diverse yet functionally valid schematics, showing that the model performs genuine reasoning rather than relying on fixed templates. Overall, this approach demonstrates strong potential for scaling schematic generation and optimization to more complex analog and mixed-signal systems.
The source code is available on https://github.com/eelab-dev.

\section*{Acknowledgment}

The authors thank EDINA and ISG@University of Edinburgh for their support in accessing OpenAI services. Also, Google Inc. for providing access to Gemini's AI services. We also thank Strategic Blue for enabling and advising on cloud-based AI services.

\bibliographystyle{IEEEtran}
\bibliography{IEEEabrv,bib}

\vspace{12pt}

\end{document}